\documentclass[letterpaper]{article}
\usepackage{ijcai13}
\usepackage{times}
\usepackage{amsmath}
\usepackage{graphicx}
\usepackage{named}

\long\def\BOC#1\EOC{\message{(Commented text )}}
\long\def\BOCC#1\EOCC{\message{(Commented text )}}
\long\def\BOCCC#1\EOCCC{\message{(Commented text )}}
\long\def\optional#1{\empty}

\def\ar{\leftarrow}
\def\beq{\begin{equation}}
\def\eeq#1{\label{#1}\end{equation}}
\def\ba{\begin{array}}
\def\ea{\end{array}}
\def\bi{\begin{itemize}}
\def\ei{\end{itemize}}
\def\i#1{\hbox{\it #1\/}}
\def\mi#1{\mathit{#1}}
\def\mu#1{\mathit{\underline{#1}}}

\def\sm{\hbox{\rm SM}}

\def\fsm{\hbox{\rm SM}}
\def\no{\i{not}}
\def\rar{\rightarrow}
\def\lrar{\leftrightarrow}

\def\v{\widehat}
\def\bC{{\bf{c}}}
\def\vbC{{\v{\bf{c}}}}

\def\false{\hbox{\sc false}}
\def\true{\hbox{\sc true}}

\def\mvis{\!=\!}
\def\sneg{\sim\!\!}

\newtheorem{thm}{Theorem}

\newtheorem{example}{Example}

\title{Answer Set Programming Modulo Theories and \\
  Reasoning about Continuous Changes}

\author{Joohyung Lee and Yunsong Meng \\
School of Computing, Informatics and Decision Systems Engineering \\
Arizona State University, Tempe, USA \\
{\tt \{joolee, Yunsong.Meng\}@asu.edu}
}

\begin{document}

\maketitle

\begin{abstract}
{\sl Answer Set Programming Modulo Theories (ASPMT)} is a new framework of
tight integration of answer set programming (ASP) and satisfiability
modulo theories (SMT). Similar to the relationship between first-order
logic and SMT, it is based on a recent proposal of the functional
stable model semantics by fixing interpretations of background
theories. Analogously to a known relationship between ASP and SAT, 
``tight'' ASPMT programs can be translated into SMT instances. We
demonstrate the usefulness of ASPMT by enhancing action language
${\cal C}$+ to handle continuous changes as well as discrete
changes. We reformulate the semantics of ${\cal C}$+ in terms of
ASPMT, and show that SMT solvers can be used to compute the
language. We also show how the language can represent cumulative
effects on continuous resources. 
\end{abstract}

\section{Introduction} \label{sec:intro}

The success of answer set programming (ASP)~\cite{lif08,bre11} is in
part thanks to efficiency of ASP solvers that utilize (i) intelligent
grounding---the process that replaces variables with ground
terms---and (ii) efficient search methods that originated from
propositional satisfiability solvers (SAT solvers).
However, this method is not scalable when we need to represent 
functional fluents that range over large numeric domains. 
Since answer sets (a.k.a. stable models) are limited to Herbrand
models, {\em functional} fluents are represented by \emph{predicates},
but not by \emph{functions}, as in the following example that
represents the inertia assumption on $\i{Speed}$:
\beq
\ba l
  \i{Speed}_1(x)\ \ar\  \i{Speed}_0(x),\  \no\ \neg
\i{Speed}_1(x)\ .
\ea
\eeq{speed-asp}
Here the subscripts $0$ and $1$ are time stamps, and $x$ is a
variable ranging over the value domain of fluent $\i{Speed}$. 
A more natural representation using functions which replaces
$\i{Speed}_i(x)$ with $\i{Speed}_i\mvis x$ does not work under
the ASP semantics because (i) answer sets are Herbrand
models (e.g., $\i{Speed}_1 = \i{Speed}_0$ is
always false under any Herbrand interpretation), and (ii) 
nonmonotonicity of the stable model semantics has to do with
minimizing the extents of predicates but has nothing to do with
functions. On the other hand, grounding rule~\eqref{speed-asp} yields
a large set of ground instances when $\i{Speed}$ ranges over a large
integer domain. Moreover, a real number domain is not supported 
because grounding cannot be applied.

In order to alleviate this ``grounding problem,'' there have been
several recent efforts to integrate ASP with constraint solving and 
satisfiability modulo theories (SMT), where functional fluents can be
expressed by variables in Constraint Satisfaction Problems or
uninterpreted constants in SMT. 
\citeauthor{balduccini09representing}~[\citeyear{balduccini09representing}]
and \citeauthor{gebser09constraint}~[\citeyear{gebser09constraint}]
combine ASP and constraint solving in a way that is similar to the
``lazy approach'' of SMT
solvers. \citeauthor{niemela08stable}~[\citeyear{niemela08stable}]
shows that ASP programs can be translated into a particular SMT
fragment, namely difference logic. 
\citeauthor{janhunen11tight}~[\citeyear{janhunen11tight}]
presents a tight integration of ASP and SMT. All these approaches
aim at addressing the issue (i) above by avoiding grounding and
exploiting efficient constraint solving methods applied to functions.
However, they do not address the
issue (ii) because, like the standard ASP, nonmonotonicity of those
extensions has to do with predicates only.
For instance, a natural counterpart of~\eqref{speed-asp} in
the language of {\sc clingcon}~\cite{gebser09constraint},
\[
  \i{Speed}_1 =^\$ x\ \ar\ \i{Speed}_0 =^\$ x,\
             \no\ \i{Speed}_1\ne^\$ x\ ,
\]
does not correctly represent inertia.

In this paper, we present a framework of combining answer set
programming with satisfiability modulo theories, which we call
{\em Answer Set Programming Modulo Theories (ASPMT)}.
Just as SMT is a generalization of SAT and, at the same time, a
special case of first-order logic in which certain predicate and
function constants in background theories (such as difference logic
and the theory of reals) have fixed interpretations, ASPMT is a
generalization of the standard ASP and, at the same time, a special
case of the functional stable model semantics
\cite{bartholomew12stable} that assumes background theories.
Like the known relationship between SAT and ASP that {\sl tight} ASP
programs can be translated into SAT instances, {\sl tight} ASPMT
programs can be translated into SMT instances, which allows SMT
solvers for computing ASPMT programs.

These results allow us to enhance action language ${\cal
  C}$+ \cite{giu04} to handle reasoning about continuous changes.
Language ${\cal C}$+ is an expressive action description language but
its semantics was defined in terms of propositional causal theories,
which limits the language to express discrete changes
only. By reformulating ${\cal C}$+ in terms of ASPMT, we naturally
extend the language to overcome the limitation, and use SMT solvers to
compute the language. Our experiments show that
this approach outperforms the existing implementations of ${\cal C}$+
by several orders of magnitude for some benchmark problems.

Section~\ref{sec:prelim} reviews the functional stable model semantics by
\citeauthor{bartholomew12stable}~[\citeyear{bartholomew12stable}], 
and defines ASPMT as its special case. Section~\ref{sec:cplus}
reformulates language ${\cal C}$+ in terms of ASPMT, and
Section~\ref{sec:continuous} shows how the reformulation can be used
to represent continuous changes. The language is further extended in
Section~\ref{sec:additive} to represent cumulative effects on
continuous resources. Related work and the conclusion are presented in
Section~\ref{sec:related}.

\section{Functional SM and ASPMT} \label{sec:prelim}

\subsection{Review: Functional Stable Model Semantics}

We review the semantics from~\cite{bartholomew12stable}.
Formulas are built the same as in first-order logic. A signature
consists of {\em function constants} and {\em predicate constants}.
Function constants of arity $0$ are called {\em object constants}, and
predicate constants of arity $0$ are called {\em propositional
  constants}.

Similar to circumscription, for predicate symbols (constants or
variables) $u$ and $c$, expression $u\le c$ is defined as shorthand for
\hbox{$\forall {\bf x}(u({\bf x})\rar c({\bf x}))$}.
Expression $u=c$ is defined as
\hbox{$\forall {\bf x}(u({\bf x})\lrar c({\bf x}))$}
if $u$ and $c$ are predicate symbols, and
$\forall {\bf x}(u({\bf x})=c({\bf x}))$
if they are function symbols.
For lists of symbols ${\bf u}=(u_1,\dots,u_n)$ and
${\bf c}=(c_1,\dots,c_n)$, expression ${\bf u}\le {\bf c}$ is defined
as $(u_1\le c_1)\land\dots\land (u_n\le c_n)$,
and similarly, expression ${\bf u} = {\bf c}$ is defined
as $(u_1 = c_1)\land\dots\land (u_n = c_n)$.

Let $\bC$ be a list of distinct predicate and function constants, and
let $\vbC$ be a list of distinct predicate and function variables
corresponding to~$\bC$.
By $\bC^\mi{pred}$ ($\bC^\mi{func}$, respectively) we mean the
list of all predicate constants (function constants, respectively) in
$\bC$, and by $\vbC^\mi{pred}$ the list of the corresponding predicate
variables in $\vbC$.

For any formula $F$, expression $\fsm[F; \bC]$ is defined as
\[
   F\land\neg\exists \vbC(\vbC<\bC\land F^*(\vbC)),
\]
where $\vbC<\bC$ is shorthand for
$(\vbC^\mi{pred}\le \bC^\mi{pred})\land\neg (\vbC=\bC)$,
and $F^*(\vbC)$ is defined recursively as follows.
\begin{itemize}\addtolength{\itemsep}{-0.3mm}
\item When $F$ is an atomic formula, $F^*$ is $F'\land F$ where $F'$
  is obtained from $F$ by replacing all intensional (function and
  predicate) constants ${\bf c}$ in it with the corresponding
  (function and predicate) variables from $\v{\bC}$;

\item  $(G\land H)^* = G^*\land H^*$;\ \ \ \ \
       $(G\lor H)^* = G^*\lor H^*$;

\item  $(G\rar H)^* = (G^*\rar H^*)\land (G\rar H)$;

\item  $(\forall x G)^* = \forall x G^*$;\ \ \ \ \
       $(\exists xF)^* = \exists x F^*$.
\end{itemize}
(We understand $\neg F$ as shorthand for $F\rar\bot$; $\top$ as
$\neg \bot$; and $F\lrar G$ as $(F\rar G)\land (G\rar F)$.)
Members of $\bC$ are called {\sl intensional} constants.

When $F$ is a sentence, the models of $\fsm[F; {\bf c}]$ are called
the {\em stable} models of~$F$ {\em  relative to} ${\bf c}$. They are
the models of~$F$ that are ``stable'' on ${\bf c}$.
The definition can be easily extended to formulas of many-sorted
signatures.

This definition of a stable model is a proper generalization of the
one from~\cite{ferraris11stable}:  in the absence of intensional
function constants, it reduces to the one in~\cite{ferraris11stable}.

\subsection{Answer Set Programming Modulo Theories}


Formally, an SMT instance is a formula in many-sorted first-order
logic, where some designated function and predicate constants are
constrained by some fixed background interpretation. SMT is the
problem of determining whether such a formula has a model that expands
the background interpretation~\cite{barrett09satisfiability}.

The syntax of ASPMT is the same as that of SMT. Let $\sigma^{bg}$ be
the (many-sorted) signature of the background theory~$bg$. An
interpretation of $\sigma^{bg}$ is called a {\em background
  interpretation} if it satisfies the background theory. For instance,
in the theory of reals, we assume that $\sigma^{bg}$ contains the set
$\mathcal{R}$ of symbols for all real numbers, the set of arithmetic
functions over real numbers, and the set $\{<, >, \le, \ge\}$ of
binary predicates over real numbers. Background interpretations
interpret these symbols in the standard way.

Let $\sigma$ be a signature that is disjoint from $\sigma^{bg}$.
We say that an interpretation $I$ of $\sigma$ satisfies $F$
w.r.t. the background theory $bg$, denoted by $I\models_{bg} F$,
if there is a background interpretation $J$ of $\sigma^{bg}$ that has
the same universe as $I$, and $I\cup J$ satisfies $F$.
For any ASPMT sentence $F$ with background theory
$\sigma^{bg}$, interpretation $I$ is a stable model of $F$ relative
to~${\bf c}$ (w.r.t. background theory $\sigma^{bg}$) if
$I\models_{bg} \sm[F; \bC]$.

We will often write $G\ar F$, in a rule form as in logic programs, 
to denote the universal closure of $F\rar G$. 
A finite set of formulas is identified with the conjunction of the
formulas in the set.

\begin{example}\label{ex:0}
The following set $F$ of formulas describes the inertia assumption on
the speed of a car and the effect of acceleration assuming the theory of
reals as the background theory. 
\beq
\ba {rcl}
  \i{Speed}_1\mvis x &\!\!\ar\!\!& \neg\neg (\i{Speed}_1\mvis x)\land
        \i{Speed}_0\mvis x\\
  \i{Speed}_1\mvis x &\!\!\ar\!\!&
      (x = \i{Speed}_0\!+\!3\!\times\i{Duration})\\ 
  & & \land\ \i{Accelerate}\mvis\true\ .
\ea
\eeq{speed}
Here $x$ is a variable of sort $\mathcal{R}_{\ge 0}$; 
$\i{Speed}_0$, $\i{Speed}_1$ and $\i{Duration}$ are object
constants with value sort ${\cal R}_{\ge 0}$ and $\i{Accelerate}$ is
an object constant with value sort Boolean. 
For interpretation $I$ of signature
$\{\i{Speed}_0,\i{Speed}_1,\i{Duration}, \i{Accelerate}\}$ 
such that
$\i{Accelerate}^I = \false$,
$\i{Speed}_0^I= 1$,
$\i{Speed}_1^I= 1$,
$\i{Duration}^I= 1.5$,
we check that $I\models_{bg} \sm[F; \i{Speed}_1]$.

For another interpretation $I_1$ of the same signature that
agrees with $I$ except that
$\i{Accelerate}^{I_1} = \true$, $\i{Speed}_1^{I_1}=5.5$,
we check that $I_1\models_{bg} \sm[F;\i{Speed}_1]$.

\end{example}

\subsection{Review: Completion}  \label{ssec:completion}

We review the {\em theorem on completion}
from~\cite{bartholomew13functional-this}, which 
turns the functional stable model semantics into first-order logic. 

A formula $F$ is said to be in {\em Clark normal form} (relative to
the list ${\bf c}$ of intensional constants) if it is a conjunction of
sentences of the form 
\beq
   \forall {\bf x} (G\rar p({\bf x}))
\eeq{cnf-p}
and
\beq
\forall {\bf x}y (G \rar f({\bf x}) \mvis y)
\eeq{cnf-f}
one for each intensional predicate $p$ and each intensional function
$f$, where ${\bf x}$ is a list of distinct object variables, $y$ is
an object variable, and $G$ is an arbitrary formula that has no free variables
other than those in ${\bf x}$ and $y$.

The {\em completion} of a formula $F$ in Clark normal form (relative
to ${\bf c}$) is obtained from $F$ by replacing each conjunctive term
(\ref{cnf-p}) with
$\forall {\bf x} (p({\bf x})\lrar G)$
and each conjunctive term (\ref{cnf-f}) with
$\forall {\bf x}y (f({\bf x})\mvis y\lrar G)$.

An occurrence of a symbol or a subformula in a formula
$F$ is called {\em strictly positive} in $F$ if that occurrence is not
in the antecedent of any implication in $F$.
The {\em t-dependency graph} of $F$ (relative to ${\bf c}$) is the
directed graph that
\begin{itemize}\addtolength{\itemsep}{-0.3mm}
\item  has all members of ${\bf c}$ as its vertices, and
\item  has an edge from $c$ to $d$ if, for some strictly positive
  occurrence of $G\rar H$ in~$F$,
  \begin{itemize}\addtolength{\itemsep}{-0.3mm}
  \item  $c$ has a strictly positive occurrence in~$H$, and
  \item  $d$ has a strictly positive occurrence in~$G$.
  \end{itemize}
\end{itemize}

We say that $F$ is {\em tight} (on {\bf c}) if the t-dependency graph of
$F$ (relative to {\bf c}) is acyclic. For example,
formula~\eqref{speed-asp} is tight on its signature.

%
\begin{thm}~\cite[Theorem 2]{bartholomew13functional-this} \label{thm:completion}\optional{thm:completion}
For any sentence~$F$ in Clark normal form that is tight on ${\bf c}$,
an interpretation $I$ that satisfies $\exists xy(x \ne y)$ is a model
of $\sm[F;{\bf c}]$ iff $I$ is a model of the completion of $F$
relative to~${\bf c}$.
\end{thm}


Theorem~\ref{thm:completion} can be applied to formulas in non-Clark
normal form if they can be rewritten in Clark normal form. 
The following theorem is often useful in splitting conjunctive
formulas, and applying completion to subformulas. 
\begin{thm}~\cite[Theorem 1]{bartholomew12stable}\label{thm:negative}
For any first-order formulas~$F$ and~$G$, if $G$ has no strictly
positive occurrence of a constant from $\bC$, \hbox{$\sm[F\land G;
  \bC]$} is equivalent to~$\sm[F; \bC]\land G$.
\end{thm}

Theorem~\ref{thm:completion} is applicable to ASPMT formulas as well.
Since $F$ in Example~\ref{ex:0} is tight on $\i{Speed}_1$,
according to Theorem~\ref{thm:completion}, $\sm[F; \i{Speed}_1]$ 
is equivalent to the following SMT instance with the same background
theory: 
\[
\ba l
  \i{Speed}_1\mvis x\ \lrar\
     \neg\neg(\i{Speed}_1\mvis x)\land (\i{Speed}_0\mvis x) \\
\hspace{0.8cm}
     \lor\ ((x\mvis \i{Speed}_0\!+\!3\!\times\i{Duration})\land
            \i{Accelerate}\mvis\true)
     \ .
\ea
\]

\section{Reformulating ${\cal C}$+ in ASPMT}\label{sec:cplus}

\subsection{Syntax}

We consider a many-sorted first-order signature $\sigma$ that is
partitioned into three signatures: the set $\sigma^{fl}$ of object
constants called {\em fluent constants}, the set $\sigma^{act}$ of
object constants called {\em action constants}, and the background
signature~$\sigma^{bg}$. The signature $\sigma^{fl}$ is further
partitioned into the set $\sigma^{sim}$ of {\em simple} fluent
constants and the set $\sigma^{sd}$ of {\em statically determined}
fluent constants.

We assume the same syntax of formulas as in Section~\ref{sec:prelim}.
A {\em fluent formula} is a formula of
signature~$\sigma^{fl}\cup\sigma^{bg}$. An {\em action formula} is a
formula of~$\sigma^{act}\cup\sigma^{bg}$ that contains at least one
action constant and no fluent constants.

A {\em static law} is an expression of the form
\beq
   {\bf caused}\ F\ {\bf if}\ G
\eeq{static}
where $F$ and $G$ are fluent formulas.
An {\em action dynamic law} is an expression of the form
(\ref{static}) in which $F$ is an action formula and $G$ is a
formula.
A {\em fluent dynamic law} is an expression of the form
\beq
{\bf caused}\ F\ {\bf if}\ G\ {\bf after}\ H
\eeq{dynamic}
where~$F$ and~$G$ are fluent formulas and $H$ is a formula, provided that~$F$
does not contain statically determined constants.
A {\sl causal law} is a static law, or an action dynamic law, or a fluent
dynamic law.
A {\em ${\cal C}$+ action description} is a finite set of causal laws.

For any function constant $f$, we say that a first-order formula is
{\em $f$-plain} if each atomic formula in it
\begin{itemize}\addtolength{\itemsep}{-0.5mm}
\item  does not contain $f$, or
\item  is of the form $f({\bf t}) = t_1$ where ${\bf t}$ is a list of
  terms not containing $f$, and $t_1$ is a term not containing~$f$.
\end{itemize}
For any list $\bC$ of predicate and function constants, we say that $F$
is $\bC$-plain if $F$ is $f$-plain for each function constant $f$ in
$\bC$.

We call an action description {\em definite} if the head $F$ of every
causal law \eqref{static} and \eqref{dynamic} is an atomic formula
that is $(\sigma^{fl}\cup\sigma^{act})$-plain. Throughout this paper we
consider definite action descriptions only, which covers the
fragment of ${\cal C}$+ that is implemented in the Causal Calculator
({\sc CCalc})\footnote{
http://www.cs.utexas.edu/~tag/ccalc/}.

\subsection{Semantics}\label{ssec:semantics}

In~\cite{giu04} the semantics of ${\cal C}$+ is defined in terms of
nonmonotonic propositional causal theories, in which every constant
has a finite domain. 
The semantics of the enhanced ${\cal C}$+ below is similar to the one
in~\cite{giu04} except that it is defined in terms of ASPMT in place
of causal theories. 
This reformulation is essential for the language to represent
continuous changes as it is not limited to finite domains only.


For a signature $\sigma$ and a nonnegative integer $i$, expression
\hbox{$i:\sigma$} is the signature consisting of the pairs $i:c$ such
that \hbox{$c\in \sigma$}, and the value sort of $i:c$ is the same as
the value sort of $c$. Similarly, if $s$ is an interpretation of
$\sigma$, expression $i:s$ is an interpretation of $i:\sigma$ such
that $c^s = (i:c)^{i:s}$.

For any definite action description $D$ of signature
\hbox{$\sigma^{fl}\cup\sigma^{act}\cup\sigma^{bg}$} and any nonnegative
integer $m$, the ASPMT program $D_m$ is defined as follows. The
signature of $D_m$ is 
$0\!:\!\sigma^{fl}\cup\dots\cup
m\!:\!\sigma^{fl}\cup0\!:\!\sigma^{act}\cup\dots\cup
(m\!-\!1)\!:\!\sigma^{act}\cup\sigma^{bg}$.
By $i:F$ we denote the result of inserting $i:$ in front of every
occurrence of every fluent and action constant in a formula~$F$.

ASPMT program $D_m$ is the conjunction of
\[
 \neg\neg\ i\!:\!G \rar i\!:\!F
\]
for every static law \eqref{static} in $D$ and every
$i\in\{0,\dots,m\}$, and for every action dynamic law \eqref{static}
in $D$ and every \hbox{$i\in\{0,\dots,m\!-\!1\}$};
\[
  \neg\neg\ (i\!+\!1)\!:\! G \land i\!:\!H \rar (i\!+\!1)\!:\!F
\]
for every fluent dynamic law \eqref{dynamic} in $D$ and every
$i\in\{0,\dots, m-1\}$.

The transition system represented by an action description~$D$
consists of states (vertices) and transitions (edges).
A {\em state} is an interpretation $s$ of $\sigma^{fl}$ such that
\hbox{$0\!:\!s\models_{bg} \sm[D_0;\ 0\!\!:\!\!\sigma^{sd}]$}.
A {\em transition} is a triple $\langle s,e,s'\rangle$, where $s$ and
$s'$ are interpretations of $\sigma^{fl}$ and $e$ is an interpretation
of $\sigma^{act}$, such that
\[
\ba l
  (0\!:\! s)\cup (0\!:\!e) \cup (1\!:\!s')\models_{bg} \\
\hspace{2.6cm} \sm[D_1;\ (0\!:\!\sigma^{sd})\cup (0\!:\!\sigma^{act})\cup
  (1\!:\!\sigma^{fl})]\ . 
\ea
\]

Theorems~\ref{thm:state} and \ref{thm:transition} below extend 
Propositions 7 and 8 from~\cite{giu04} to our reformulation of~${\cal C}$+. 
These theorems justify the soundness of the language. 

The definition of the transition system above implicitly relies on the
following property of transitions:

\begin{thm}\label{thm:state}\optional{thm:state}
For every transition $\langle s,e,s'\rangle$, $s$ and $s'$ are states.
\end{thm}

The following theorem states the correspondence between the stable
models of $D_m$ and the paths in the transition system represented by
$D$:

\begin{thm}\label{thm:transition}\optional{thm:transition}
\[
\ba l
(0\!:\!s_0) \cup (0\!:\!e_0) \cup (1\!:\!s_1) \cup (1\!:\!e_1) \cup
        \cdots \cup (m\!:\!s_m) \\
\hspace{0.2cm}\models_{bg}
\sm[D_m;\ (0\!:\!\sigma^{sd})\cup
         (0\!:\!\sigma^{act})\cup
         (1\!:\!\sigma^{fl})\cup
         (1\!:\!\sigma^{act})  \\
\hspace{3.5cm}  \cup\dots\cup
        (m\!-\!1\!:\!\sigma^{act})\cup
        (m\!:\!\sigma^{fl})]
\ea
\]
iff each triple $\langle s_i,e_i,s_{i+1}\rangle$ $(0\le i<m)$ is a
transition.
\end{thm}


It is not difficult to check that ASPMT program $D_m$ that is
obtained from action description $D$ is always tight. In view of Theorem~\ref{thm:completion},  $D_m$ can be
represented in the language of SMT as the next section demonstrates.

\section{Reasoning about Continuous Changes in ${\cal C}$+}\label{sec:continuous}

In order to represent continuous changes in the enhanced ${\cal C}$+,
we distinguish between steps and real clock times. We assume the
theory of reals as the background theory, and introduce a simple
fluent constant $\i{Time}$ with value sort $\mathcal{R}_{\ge 0}$, which
denotes the clock time, and an action constant $\i{Dur}$ with value sort
$\mathcal{R}_{\ge 0}$, which denotes the time elapsed between the two
consecutive states. We postulate the following causal laws:
\beq
\ba l
  {\bf exogenous}\ \i{Time},  \i{Dur}, \\
  {\bf constraint}\ \i{Time}\mvis t+t'\
     {\bf after}\ \i{Time}\mvis t\land\i{Dur}\mvis t' \ .
\ea
\eeq{time-progress}
These causal laws are shorthand for
\[
\ba l
{\bf caused}\ \i{Time}\mvis t\ {\bf if}\ \i{Time}\mvis t\ , \\
{\bf caused}\ \i{Dur}\mvis t\ {\bf if}\ \i{Dur}\mvis t\ , \\
{\bf caused}\ \bot\ {\bf if}\ \neg (\i{Time}\mvis t+t')\
  {\bf after}\ \i{Time}\mvis t\land\i{Dur}\mvis t' \
\ea
\]
where $t, t'$ are variables of sort $\mathcal{R}_{\ge 0}$. (See
Appendix~B in~\cite{giu04} for the abbreviations of causal laws.)

Continuous changes can be described as a function of duration using 
fluent dynamic laws of the form
\[
\small
\ba l
  {\bf caused}\ c\mvis f({\bf x}, {\bf x}', t)\
  {\bf if}\ {\bf c}'\mvis{\bf x}'\ 
  {\bf after}\ ({\bf c}\mvis{\bf x})\land(\i{Dur}\mvis t)\land G
\ea
\]
where (i) $c$ is a simple fluent constant, (ii) ${\bf c}$, ${\bf c}'$ are lists
of fluent constants,
(iii) ${\bf x}$, ${\bf x}'$ are lists of object variables,
(iv) $G$ is a formula, and
(v) $f({\bf x}, {\bf x}', t)$ is a term constructed
from~$\sigma^{bg}$, and variables in ${\bf x}$, ${\bf x}'$, and $t$.

For instance, the fluent dynamic law
\[ 
\ba l
  {\bf caused}\ \i{Distance}\mvis
        d\!+\!0.5\!\times\! (v\!+\!v')\!\times\! t\
  {\bf if}\ \i{Speed}\mvis v'\\
\hspace{0.5cm} {\bf after}\ \i{Distance}\mvis d
     \land\i{Speed}\mvis v
     \land\i{Dur}\mvis t
\ea
\] 
describes how fluent $\i{Distance}$ changes according to the function
of real time.

Consider the following problem by
Lifschitz
({http://www.cs.utexas.edu/vl/tag/continuous\_problem}).
If the accelerator of a car is activated, the
car will speed up with constant acceleration ${\rm A}$ until the
accelerator is released or the car reaches its maximum speed
${\rm MS}$, whichever comes first. If the brake is activated, the car
will slow down with acceleration $-{\rm A}$ until the brake is
released or the car stops, whichever comes first. Otherwise, the speed
of the car remains constant. The problem asks to find a plan
satisfying the following condition: at time 0, the car is at rest at
one end of the road; at time ${\rm K}$, it should be at rest at the other
end.



\begin{figure}[t]
{\footnotesize
\hrule
\smallskip
\begin{tabbing}
Notation: $d$, $v$, $v'$, $t$, $t'$ are variables of sort $\mathcal{R}_{\ge 0}$\\ 
{\rm A}, {\rm MS} are real numbers. \\  \\
Simple fluent constants:   \hskip 2cm  \= Domains:\\
~~~~$\i{Speed}$, $\i{Distance}$, $\i{Time}$   \> ~~~~$\mathcal{R}_{\ge 0}$\\
Action constants:                      \> Domains:\\
~~~~$\i{Accelerate}$, $\i{Decelerate}$       \> ~~~~Boolean\\
~~~~$\i{Dur}$                             \>~~~~$\mathcal{R}_{\ge 0}$\\  \\


${\bf caused}\ \i{Speed}\mvis v\!+\!{\rm A}\!\times\! t\
  {\bf after}\ \i{Accelerate}\land
   \i{Speed}\mvis v\land\i{Dur}\mvis t$ \\

${\bf caused}\ \i{Speed}\mvis v\!-\!{\rm A}\!\times\! t\
   {\bf after}\ \i{Decelerate}\land
   \i{Speed}\mvis v\land\i{Dur}\mvis t$ \\

${\bf caused}\  \i{Distance}\mvis
   d\!+\!0.5\!\times\!(v\!+\!v')\!\times\! t\
   {\bf if}\ \i{Speed}\mvis v'$ \\
\hspace{3cm} ${\bf after}\ 
      \i{Distance}\mvis d\land
      \i{Speed}\mvis v\land
      \i{Dur}\mvis t$ \\

${\bf constraint}\ \i{Time}\mvis t\!+\!t'\
  {\bf after}\ \i{Time}\mvis t\land \i{Dur}\mvis t'$ \\

${\bf constraint}\ \i{Speed}\le{\rm MS}$ \\ \\

${\bf inertial}\ \i{Speed}$ \\

${\bf exogenous}\ \i{Time}, c$ \hskip 2cm \= for every action
constant~$c$
\end{tabbing}
\hrule
}
\caption{{\small Car Example in ${\cal C}$+}}
\vspace{-4mm}
\label{fig:car-cplus}
\end{figure}

\begin{figure}[b]
{\footnotesize
\hrule
\smallskip
\begin{tabbing}
Notation: $x$, $d$, $v$, $v'$, $t$, $t'$ are variables of sort $\mathcal{R}_{\ge 0}$; \\
\hspace{1.2cm} $y$ is a variable of sort Boolean. \\

Intensional object constants:  $i\!:\!\i{Speed}$ for $i>0$ \\ \\
$i\!+\!1\!:\!\i{Speed}\mvis x\ \ar\
   (x = v\!+\!{\rm A}\!\times\! t)$ \\
   \hspace{3cm} $\land\ i\!:\!(\i{Accelerate}\land\i{Speed}\mvis v\land\i{Dur}\mvis t)$ \\

$i\!+\!1\!:\!\i{Speed}\mvis x\ \ar\
   (x = v\!-\!{\rm A}\!\times\! t)$ \\
\hspace{3cm} $\land\ i\!:\!(\i{Decelerate}\land\i{Speed}\mvis
v\land\i{Dur}\mvis t)$ \\

$i\!+\!1\!:\!\i{Distance}\mvis x\ \ar\
  (x =  d\!+\!0.5\!\times\!(v'\!+\!v)\!\times\! t)$ \\  
\hspace{0.5cm} $\land\ i\!+\!1\!:\!\i{Speed}\mvis v'\land
     i:(\i{Speed}\mvis v\land
     \i{Distance}\mvis d\land\i{Dur}\mvis t)$ \\


$\bot\ar\neg (i\!+\!1\!:\!\i{Time}\mvis t\!+\!t')\land 
  i:(\i{Time}\mvis t\land \i{Dur}\mvis t')$ \\

$\bot\ar\neg (i\!:\!\i{Speed}\le {\rm MS})$ \\

$i\!+\!1\!:\!\i{Speed}\mvis x \ar \neg\neg (i\!+\!1\!:\!\i{Speed}\mvis
x) \land i\!:\!\i{Speed}\mvis x$ \\

$i\!:\!\i{Time}\mvis t\ar\neg\neg (i\!:\!\i{Time}\mvis t)$ \\

$i\!:\!c\mvis y\ar\neg\neg (i\!:\!c\mvis y)$ \hskip 1cm for every action constant
$c$
\end{tabbing}
\hrule
}
\caption{{\small Car Example in ASPMT}}
\label{fig:car-aspmt}
\end{figure}

A ${\cal C}$+ description of this example is shown in
Figure~\ref{fig:car-cplus}. 
The actions $\i{Accelerate}$ and
  $\i{Decelerate}$ has direct effects on $\i{Speed}$ and indirect
  effects on $\i{Distance}$. 
For an object constant $c$ that has the Boolean domain, we abbreviate
$c\mvis\true$ as $c$ and $c\mvis\false$ as $\sneg c$.
According to the semantics in
Section~\ref{ssec:semantics}, the description is turned into
an ASPMT program with the theory of reals as the background theory,
which can be further rewritten in Clark normal form. Some occurrences
of $\neg\neg$ can be dropped without affecting stable models, which
results in the program in Figure~\ref{fig:car-aspmt}.

The program can be viewed as $F\land G$ where $F$ is the conjunction
of the rules that has intensional constants in the heads, and $G$ is
the conjunction of the rest rules. In view of
Theorem~\ref{thm:negative}, the stable models and $F\land G$ are the
same as the stable models of $F$ that satisfies $G$. 
By Theorem~\ref{thm:completion}, $F$ can be turned into completion. 
For example, the completion on $i\!+\!1\!:\!\i{Speed}$ yields a
formula that is equivalent to 
\[ 
\small
\ba l
i\!+\!1\!:\!\i{Speed}\mvis x\lrar
\hspace{0cm} \big(x = (i\!:\!\i{Speed}\!+\!A\!\times\! i\!:\!\i{Dur})
   \ \land\ i\!:\!\i{Accelerate}\big) \\
\hspace{1cm} \lor \big(x = (i\!:\!\i{Speed}\!-\!A\!\times\!i\!:\!\i{Dur})
  \ \land\ i\!:\!\i{Decelerate}\big) \\
\hspace{1cm} \lor \big(i\!+\!1\!:\!\i{Speed}\mvis x\ \land\ i\!:\!\i{Speed}\mvis x\big).
\ea
\] 
Variable $x$ in the formula can be eliminated by equivalent
transformations using equality: 
\[
\small
\ba l
  i\!:\!\i{Accelerate}\rar
  i\!+\!1\!:\!\i{Speed}\mvis (i\!:\!\i{Speed}\!+\!A\!\times\!
  i\!:\!\i{Dur})\\

  i\!:\!\i{Decelerate} \rar
  i\!+\!1\!:\!\i{Speed}\mvis (i\!:\!\i{Speed}\!-\!A\!\times\!
  i\!:\!\i{Dur}) \\

  (i\!+\!1:\i{Speed}= (i\!:\!\i{Speed}\!+\!A\!\times\!
  i\!:\!\i{Dur})\land i\!:\!\i{Accelerate})  \\
  \hspace{0.5cm}\lor(i\!+\!1:\i{Speed}= (i\!:\!\i{Speed}\!-\!A\!\times\!
  i\!:\!\i{Dur})\land i\!:\!\i{Decelerate})  \\
  \hspace{0.5cm}\lor(i\!:\!\i{Speed}=i\!+\!1\!:\!\i{Speed})\ .
\ea
\]   

\begin{figure}[t]
\centering
\includegraphics [width=8cm]{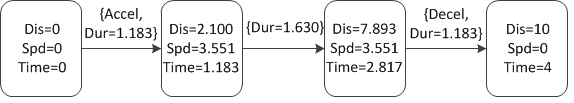}
\caption{{\small Solution Found by iSAT}}
\label{fig:car-plan}
\end{figure}

The whole translation can be encoded in the input language of
SMT solvers. The shortest plan found by
iSAT ({http://isat.gforge.avacs.org}) on this input 
formula when the road length is 10, ${\rm   A}=3, {\rm MS}=4, {\rm
  K}=4$ is shown in Figure~\ref{fig:car-plan}.

\section{Reasoning about Additive Continuous Resources in ${\cal
    C}+$} \label{sec:additive}

Additive fluents are fluents with numerical values such that the
effect of several concurrently executed actions on it can be computed
by adding the effects of the individual actions.
\citeauthor{lee03}~[\citeyear{lee03}] show how to describe additive
fluents in ${\cal C}$+ by understanding ``increment laws'' as
shorthand for some causal laws.  However, some additive fluents are
real-valued, and cannot be represented in the language described
in~\cite{lee03} as the language is limited to finite domains
only. This made the discussion of
additive fluents in~\cite{lee03} limited to integer domains only. For
example, the effect of firing multiple jets on the velocity of a
spacecraft is described by ``increment laws''
\[
\small
\ba l
  \i{Fire}(j)\ {\bf increments}\ \i{Vel}(ax)\ {\bf by}\
      n//\i{Mass}\ {\bf if}\ \i{Force}(j,ax)\mvis n\ .
\ea
\]
$\i{Mass}$ stands for an integer constant, and the symbol $//$ stands
for integer division;  the duration of firing action is assumed to be
$1$, and all components of the position and the velocity vectors at
any time are assumed to be integers, and even the forces applied have
to be integers. Obviously these are too strong assumptions. 

These limitations are not present in our enhanced ${\cal C}$+ and its
SMT-based computation. 
The representation in~\cite{lee03} can be straightforwardly extended to
handle continuous motions by distinguishing between steps and real
time as in the previous section. For example, we can describe
the effect that firing multiple jets has on the velocity of a
spacecraft by
\[
\ba l
  \i{Fire}(j)\ {\bf increments}\ \i{Vel}(ax)\ {\bf  by}\
    (n/\i{Mass})\times t\ \\
\hspace{2cm}  {\bf if}\ \i{Force}(j,ax)\mvis n\land
    \i{Dur}\mvis t\ 
\ea
\]

In general, additive fluent constants are simple fluent
  constants with numerical values with certain restrictions. First,
  the heads of static and fluent dynamic laws are not allowed to
  contain additive fluent constants. Second, the effects of concurrent
  execution of actions on additive fluents are expressed by {\sl
    increment laws\/}---expressions of the form 
\beq
  a\ {\bf increments}\ c\ {\bf by}\ f({\bf x},t)\ {\bf if}\ 
    ({\bf d}, \i{Dur})\mvis ({\bf x}, t)\land G 
\eeq{increment}
where
(i) $a$ is a Boolean action constant;
(ii) $c$ is an additive fluent constant;
(iii) ${\bf d}$ is a list of fluent constants, and ${\bf x}$ is a
  list of corresponding variables; 
(iv) $f({\bf x},t)$ is an arithmetic expression over ${\bf x}$ and the duration $t$; 
(v) $G$ is a formula that contains no Boolean action constants.

Similar to~\cite{lee03}, the semantics of increment laws is described
by a translation that replaces increment laws with action dynamic
laws~\eqref{static} and fluent dynamic laws~\eqref{dynamic} using additional
action constants. This translation largely repeats the one in~\cite{lee03} and
we omit the details due to lack of space.

The ${\cal C}$+ encoding of the spacecraft example is shown in
Figure~\ref{fig:spacecraft}. 

\begin{figure}[h]
{\footnotesize
\hrule
\begin{tabbing}
Notation: $j\in \{J_1, J_2\}$, $\i{ax}\in \{X, Y, Z\}$ \\
$x$, $y$, $y'$ are variables of sort $\mathcal{R}$;\ \ 
$t$, $t'$ are variables of sort $\mathcal{R}_{\ge 0}$ \\ \\

Additive fluent constants:       \hskip 2cm  \=Domains: \\
~~~~$\i{Vel}(\i{ax})$                           \>~~~~$\mathcal{R}$\\
Other Simple fluent constants:   \hskip 2cm  \> Domains:\\
~~~~$\i{Pos}(\i{ax})$   \> ~~~~$\mathcal{R}$\\
~~~~$\i{Time}$   \> ~~~~$\mathcal{R}_{\ge 0}$\\
Action constants:                      \> Domains:\\
~~~~$\i{Fire}(j)$                         \> ~~~~Boolean\\
~~~~$\i{Force}(j,\i{ax})$                 \>~~~~$\mathcal{R}$ \\ 
~~~~$\i{Dur}$                             \>~~~~$\mathcal{R}_{\ge
  0}$\\ \\

$\i{Fire}(j)\ {\bf increments}\ \i{Vel}(\i{ax})\ {\bf by}\
(x /{\rm Mass})\times t \ $\\
 \hspace{4cm}${\bf if}\ \i{Force}(j,\i{ax})=x\land \i{Dur}=t$ \\

${\bf caused}\ \i{Pos}(\i{ax})\mvis
x\!+\!(0.5\!\times\!(y\!+\!y')\!\times\! t)\ 
{\bf if}\ \i{Vel}(\i{ax})\mvis y' $\\
\hspace{2cm}${\bf after}\ \i{Pos}(\i{ax})\mvis x\land
   \i{Vel}(\i{ax})\mvis y\land\i{Dur}\mvis t$ \\

${\bf always}\ \i{Force}(j,\i{ax})\mvis 0\ \lrar\ \sneg\i{Fire}(j)$ \\

${\bf constraint}\ \i{Time}\mvis t\!+\!t'\ 
  {\bf after}\ \i{Time}\mvis t\land\i{Dur}\mvis t' $\\

${\bf exogenous}\ \i{Time}$\\
${\bf exogenous}\ c$ \hskip 2cm \= for every action
constant~$c$
\end{tabbing}
\hrule
}
\caption{{\small Spacecraft Example in ${\cal C}$+ with Additive Fluents}}
\label{fig:spacecraft}
\end{figure}

\begin{table*}[t]
\centering
\scriptsize
\begin{tabular}{|c|c|c|c|c|c|c|}

\hline  Max Step & \multicolumn{2}{|c|}{{\sc CCalc} v2.0 using {\sc
    relsat} v2.2} &
\multicolumn{2}{|c|}{{\sc cplus2asp} v1.0 using {\sc gringo} v3.0.3+{\sc
    clasp} v2.0.2}& \multicolumn{2}{|c|}{${\cal C}$+ in iSAT v1.0}\\
\hline & Run Time          & \# of atoms / clauses  & Run Time          & \# of atoms / rules & Run Time & \# of variables / clauses\\
       & (grounding+completion+solving) &             & (grounding+solving) &      &    last/total & (bool + real) ~~~~~~~~~~~  \\
\hline
\hline  1   & 0.16\ (0.12+0.04+0.00)          & 488  / 1872        & 0.005\ (0.005+0)      & 1864  / 2626     & 0/0       &  (42+53) / 182       \\
\hline  2   & 0.57\ (0.40+0.17+0.00)          & 3262  / 14238        &0.033\ (0.033+0)      & 6673  / 12035    & 0/0       &  (82+98) / 352      \\
\hline  3   & 10.2\ (2.62+1.58+6)          & 32772  / 155058       &
0.434\ (0.234+0.2)    & 42778 / 92124    & 0/0       & (122+143) / 520       \\
\hline  4   & 505.86\ (12.94+13.92+479)       & 204230  / 992838
&  12.546\ (3.176+9.37)   & 228575/ 503141   & 0/0       & (162+188) /  688         \\
\hline  5   & failed\ (51.10+115.58+ failed)        & 897016 / 4410186
&  73.066\ (15.846+57.22) & 949240/ 2060834  & 0/0.03    & (202+233)  / 856         \\
\hline  6   & time out        & --           &3020.851\ (62.381+2958.47) & 3179869/ 6790167  & 0/0.03    & (242+278) /  1024    \\
\hline  10  & time out        & --           &   time out    & --          & 0.03/0.09 & (402+458)  / 1696    \\
\hline  50  & time out        & --           &   time out    & --          & 0.09/1.39 & (2002+2258) / 8416    \\
\hline  100 & time out        & --           &   time out    & --          & 0.17/5.21 & (4002+4508) /  16816 \\
\hline  200 & time out        & --           &   time out    & --          & 0.33/21.96& (8002+9008) /  33616  \\

\hline
\end{tabular}
\caption{{\small Experiment results on Spacecraft Example (time out after 2
   hours)}}
 \vspace{-4mm}
\label{tab:spacecraft}
\end{table*}

Table~\ref{tab:spacecraft} compares the performance of SMT-based
computation of ${\cal C}$+ and existing implementations of ${\cal
  C}$+: {\sc CCalc} 2 and {\sc cplus2asp}. System {\sc cplus2asp}
translates ${\cal C}$+ into ASP
programs and use {\sc gringo} and {\sc clasp} for computation.
For the sake of comparison, we assume that the duration
of each action is exactly 1 unit of time so that the plans found by
the systems are of the same kind.
We assume that initially the spacecraft is rest at coordinate
$(0,0,0)$. The task is to find a plan such that at each integer
time $t$, the spacecraft is at $(t^2,t^2,t^2)$.
The experiment was performed on an Intel Core 2 Duo CPU 3.00 GHz with 4
GB RAM running on Ubuntu version 11.10. It shows a clear advantage of
the SMT-based computation of ${\cal C}$+ for this example.


\begin{figure}[h]
{\footnotesize
\hrule
\smallskip
\begin{tabbing}
Notation: $x\in \{\i{Tap}_1, \i{Tap}_2\}$;\ \
${\rm W}(x)$, ${\rm V}$, ${\rm Low}$, ${\rm High}$ are real numbers;  \\
$t$, $t'$ are variables of sort $\mathcal{R}_{\ge 0}$. \\ \\





$\i{TurnOn}(x)\ {\bf causes}\ \i{On}(x)\land\i{Dur}\mvis 0$ \\ 
$\i{TurnOff}(x)\ {\bf causes}\ \sneg\i{On}(x)\land\i{Dur}\mvis 0$ \\ \\


$\i{On}(x)\ {\bf increments}\ \i{Level}\
   {\bf by}\ {\rm W}(x)\!\times\! t\ {\bf if}\ \i{Dur}\mvis t$\\

$\i{Leaking}\ {\bf increments}\ \i{Level}\
   {\bf by}\ -\!({\rm V}\!\times\! t)\ {\bf if}\ \i{Dur}\mvis t$ \\ \\

${\bf constraint}\ ({\rm Low}\!\le\!\i{Level})\land (\i{Level}\!\le\!{\rm High})$\\

${\bf inertial}\ \i{On}(x), \i{Leaking}$ \\

${\bf exogenous}\ c$ \hskip 2cm \= for every action constant~$c$  \\ \\

${\bf exogenous}\ \i{Time}$ \\



${\bf constraint}\ \i{Time}\mvis t+t'\
  {\bf after}\ \i{Time}\mvis t\land\i{Dur}\mvis t'$
\end{tabbing}
\hrule
\caption{Two Taps Water  Tank Example ${\cal C}$+}
\label{fig:2tap}
}
\end{figure}

The language ${\cal C}$+ is flexible enough to represent the {\em
  start-process-end} model~\cite{reiter96natural,fox06modelling},
where instantaneous actions may initiate or terminate processes.
Processes run over time and have a continuous effect on numeric
fluents. They are initiated and terminated either by the direct action
or by events that are triggered. 
This model can be encoded in ${\cal C}+$ by introducing {\em process
  fluents}, which are Boolean-valued. Such a process fluent is assumed
to be inertial, and is caused to be true or false by instantaneous
events. 
Once true, the process fluent $p$ determines the changes of additive
fluents $c$ by increment laws
\[
\ba l
  p\ {\bf increments}\ c\ {\bf by}\ f({\bf x},t)\ {\bf if}\ 
  ({\bf d},\i{Dur})\mvis ({\bf x},t) \land G\ .
\ea
\] 
Here, ${\bf increments}$ laws are defined similar to \eqref{increment}
except that $p$ is a process fluent, instead of a Boolean action
constant; we require that $G$ contain no process fluents and no
Boolean action constants. 
For example,
\[ 
\small
\ba l
\i{On}(\i{Tap}_1)\ {\bf increments}\ \i{Level}\
{\bf by}\ {\rm W}(\i{Tap}_1)\!\times\! t\ {\bf if}\ \i{Dur}\mvis t
\ea
\] 
specifies that, when the process fluent $\i{On}(\i{Tap}_1)$ is true,
the process contributes to increasing the water level by ${\rm
  W}(\i{Tap}_1)$ multiplied by duration. 

Figure~\ref{fig:2tap} describes the level of a water tank that has two
taps with different flow rates and possible leaking.


\section{Related Work and Conclusion} \label{sec:related}


Besides the functional stable model semantics we considered in this
paper, there are other approaches to incorporate  ``intensional''
functions in
ASP~\cite{cabalar11functional,lifschitz12logic,balduccini12aconservative}.

Action language ${\cal H}$~\cite{chin04,chintabathina12anew} is
similar to our enhanced ${\cal C}$+ in that it can handle reasoning
about continuous changes. One notable difference is there, each state
represents an interval of time, rather than a particular timepoint.
Language ${\cal H}$ does not have action dynamic laws, and
consequently does not allow additive fluents.

The durative action model of PDDL 2.1~\cite{fox03pddl} is similar to
our ${\cal C}$+ representation in Section~\ref{sec:continuous}.
The start-process-stop model of representing continuous changes in
PDDL+~\cite{fox06modelling} is similar to the representation in
Section~\ref{sec:additive}. Our work is also related to
\cite{shin05processes}, which used SMT solvers for computing
PDDL+, and a recent work on planning modulo
theories~\cite{gregory12planning}. 

Continuous changes have also been studied in the situation calculus,
the event calculus and temporal action logics.
In~\cite{lee12reformulating,lee12reformulating1}, these action
formalisms were reformulated in ASP. We expect that the techniques we
developed in this paper can be applied for more effective computation
of these formalisms using SMT solvers.

\medskip\noindent
{\bf Acknowledgements:} 
We are grateful to Michael Bartholomew for useful discussions related
to this paper. We are also grateful to the anonymous
referees for their useful comments. This work was partially supported
by the National Science Foundation under Grant IIS-0916116 and by the
South Korea IT R\&D program MKE/KIAT 2010-TD-300404-001.

\bibliographystyle{named}

\end{document}